\documentclass[journal]{IEEEtran}

\usepackage{ifpdf}
\usepackage{cite}
\usepackage{colortbl}
\usepackage[table]{xcolor}

\usepackage{hyperref}

\usepackage{listings}

\usepackage{multirow}
\definecolor{codegreen}{rgb}{0,0.6,0}
\definecolor{codegray}{rgb}{0.5,0.5,0.5}
\definecolor{codepurple}{rgb}{0.58,0,0.82}
\definecolor{backcolour}{rgb}{0.95,0.95,0.92}

\definecolor{darkred}{rgb}{0.55, 0.0, 0.0}
\definecolor{ultramarine}{rgb}{0.07, 0.04, 0.56}
\definecolor{upforestgreen}{rgb}{0.0, 0.27, 0.13}

\lstdefinestyle{mystyle}{
  backgroundcolor=\color{backcolour},   commentstyle=\color{codegreen},
  keywordstyle=\color{magenta},
  numberstyle=\tiny\color{codegray},
  stringstyle=\color{codepurple},
  basicstyle=\ttfamily\footnotesize,
  breakatwhitespace=false,         
  breaklines=true,                 
  captionpos=b,                    
  keepspaces=true,                 
  numbers=left,                    
  numbersep=5pt,                  
  showspaces=false,                
  showstringspaces=false,
  showtabs=false,                  
  tabsize=2
}

\definecolor{greenPython}{RGB}{0,150,0}

\usepackage{dirtytalk}
\usepackage{algorithm}
\newcommand{\Angie}[1]{{\color{blue} #1}}

\usepackage{breqn}
\usepackage{soul}
\usepackage{microtype}
\usepackage{graphicx}
\usepackage{booktabs} 
\usepackage{amsmath}
\usepackage{amsfonts}
\usepackage{bbm}
\usepackage{mathtools}

\DeclarePairedDelimiter\floor{\lfloor}{\rfloor}
\DeclareMathOperator*{\argmax}{arg\,max}
\usepackage{hyperref}
\usepackage{verbatim}

\usepackage[caption=false,font=normalsize,labelfont=sf,textfont=sf]{subfig}

\ifCLASSINFOpdf
\usepackage{graphicx}
\usepackage{amsmath}
\usepackage{algorithmic}
\usepackage{amsfonts}
\usepackage{amssymb}

\graphicspath{{Figures/}}

\usepackage{array}
\usepackage{fixltx2e}

\begin{document}
\title{LaplaceNet: A Hybrid Graph-Energy Neural  Network  for Deep Semi-Supervised Classification}

\author{Philip~Sellars$^1$, Angelica I. Aviles-Rivero$^1$ and Carola-Bibiane Sch{\"o}nlieb$^1$%
\thanks{P. Sellars, Angelica I. Aviles-Rivero and  Carola-Bibiane Sch{\"o}nlieb are with the Department of Theoretical Physics and Applied Mathematics, Univeristy of Cambridge, Cambridge, UK.  {ps644,ai323,cbs31}@cam.ac.uk .}
}

\maketitle


\begin{abstract}
Semi-supervised learning has received a lot of recent attention as it alleviates the need for large amounts of labelled data which can often be expensive, requires expert knowledge and be time consuming to collect. Recent developments in deep semi-supervised classification have reached unprecedented performance and the gap between supervised and semi-supervised learning is ever-decreasing. This improvement in performance has been based on the inclusion of numerous technical tricks, strong augmentation techniques and costly optimisation schemes with multi-term loss functions. We propose a new framework, LaplaceNet, for deep semi-supervised classification that has a greatly reduced model complexity. We utilise a hybrid approach  where pseudo-labels are produced by minimising the Laplacian energy on a graph. These pseudo-labels are then used to iteratively train a neural-network backbone. Our model outperforms state-of-the-art methods for deep semi-supervised classification, over several benchmark datasets.   Furthermore, we consider the application of strong-augmentations to neural networks theoretically and justify the use of a multi-sampling approach for semi-supervised learning. We demonstrate, through rigorous experimentation, that a multi-sampling augmentation approach improves generalisation and reduces the sensitivity of the network to augmentation.
\end{abstract}


\section{Introduction}
The advent of deep learning has been key in achieving outstanding performance in several computer vision tasks including image classification~\cite{simonyan2014very,krizhevsky2012imagenet, he2016deep,hu2018squeeze,wang2017residual}, object detection e.g.~\cite{ren2015faster,girshick2015fast,redmon2016you} and image segmentation ~\cite{long2015fully, ronneberger2015u, chen2017deeplab}. Training deep learning models often relies upon access to large amounts of labelled training data. In real-world scenarios we often find that labels are scarce, expensive to collect, prone to errors (high uncertainty) and might require expert knowledge. Therefore, relying on a well-representative dataset to achieve good performance is a major limitation for the practical deployment of machine learnt methods.  These issues have motivated the development of techniques which are less reliant on labelled data.

Semi-supervised learning aims to extract information from unlabelled data, in combination with a small amount of label data, and produce results comparable to fully supervised approaches. In recent years, the developments in deep learning have motivated new directions in semi-supervised learning (SSL) for image classification. The major benefit of these new deep approaches being the ability to learn feature representations rather than rely upon hand-crafted features. In the last few years, deep SSL papers have reached unprecedented performance e.g.~\cite{sohn2020fixmatch,xie2019unsupervised}, and the gap between supervised and semi-supervised models is much smaller now that it was even five years ago, with semi-supervised methods surpassing certain supervised techniques.

What techniques have been crucial to the improved performance of deep semi-supervised methods? Although, there is no universal answer, there are several shared commonalities between the current SOTA.  The works of ~\cite{xie2019unsupervised,sohn2020fixmatch,remixmatch} demonstrated that a key factor for improving performance is the use of strong augmentations strategies such as AutoAugment~\cite{cubuk2018autoaugment}, RandAugment~\cite{cubuk2020randaugment}, Cutout~\cite{devries2017improved} and CTAugment~\cite{remixmatch}. Additionally, the use of confidence thresholding \cite{sohn2020fixmatch,simplepseudo} and temperature sharpening \cite{xie2019unsupervised,berthelot2019mixmatch} are thought to be vital in improving performance for pseudo-labeling methods. Other papers \cite{arazo2019pseudo,verma2019interpolation,berthelot2019mixmatch} have shown great improvement from using interpolating techniques such as MixUp \cite{zhang2017mixup}. Several SOTA have also promoted large batch sizes \cite{sohn2020fixmatch} with a large ratio of unlabelled to labelled data per batch.

\begin{figure*}[ht]
	\centering
    \includegraphics[width=\textwidth]{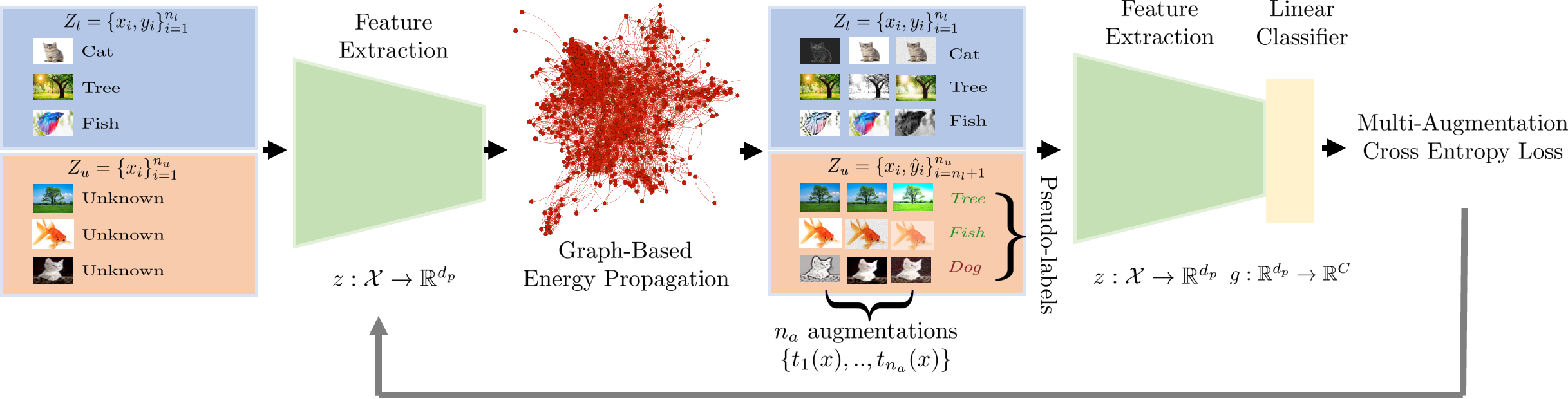}
	\caption{An overview of the approach used in LaplaceNet. 
		We start with labelled data (blue) $Z_l = \{ x_i,y_i \}_{i=1}^{n}$ and unlabelled data (orange) $Z_u = \{ x_i \}_{i=1}^{n_u}$. The data is embedded into a feature space using a neural network back-bone, which is fixed for pseudo-label generations. The features are used to construct a graphical representation of the data. We then propagate the graph Laplacian energy over the graph to obtain an estimated label for each unlabelled point, the output of the process being  the pseudo-labelled data set $Z_u = \{ x_i , \hat{y}_i \}_{i=n_l+1}^{n_u}$. The pseudo-labels $\hat{y}$ are not guaranteed to be correct, as show by the misclassification of the cat image. Each image is then augmented $n_a$ times to create multi-augment groups for each image. The full model is then trained using a simple averaged multi-augmentation cross entropy loss. The updated model is then used to create new feature embeddings and as a result more accurate pseudo-labels. This cycle of pseudo-label generation and model training continues for the duration of the algorithm.} 
	\label{fig::plv}
\end{figure*}

Recent approaches in SSL have proposed costly optimisation schemes involving multi-term loss functions to improve the generalisation of their models \cite{remixmatch,arazo2019pseudo}. Some approaches \cite{sohn2020fixmatch} use separate loss terms for unlabelled and labelled data, whilst consistency regularisation approaches such as \cite{xie2019unsupervised} use a standard supervised loss in combination with a specialised consistency loss. Other approaches go even further \cite{arazo2019pseudo,remixmatch} and use three or more loss terms which promote entropy minimisation, class balancing or simultaneously minimise several consistency losses.

Over-costly computational approaches and unnecessary complexity, make it hard to directly say what tools or approaches are important for improved generalisation and make it difficult to use SSL methods in realistic settings.  Furthermore, despite the significant improvements found in using augmentations, there has been little effort in the field of SSL to investigate how best to include strong augmentations techniques in the learning framework. With these points in mind, in this work, we introduce a new deep SSL framework for image classification which offers state-of-the-art performance with massively reduced model complexity.  Our main contributions are:

\begin{itemize}
  \item[--] We propose a graph based pseudo-label approach for semi-supervised image classification whi  ch we name LaplaceNet. We demonstrate through extensive testing, that our approach produces state-of-the-art results on benchmark datasets CIFAR-10, CIFAR-100 and Mini-ImageNet. We do so with vastly reduced model complexity compared to the current state-of-the-art. We show that a single loss, the classic supervised loss, is all that is required for fantastic performance in the SSL domain. 
  
  \item[--] We show that using an energy-based graphical model for pseudo-label generation produces more accurate pseudo-labels, with a small computational overhead, than using the network's predictions directly. Furthermore, we demonstrate that energy-based pseudo-label approaches can produce state-of-the-art results without the techniques (temperature sharpening, confidence thresholding, soft labels) that are currently thought to be essential for pseudo-label methods. 
    
  \item[--] Instead, we offer further evidence that strong augmentation is by far and away the most important tool for improving the performance of semi-supervised models in the natural image domain. With this in mind, we propose, theoretically justify and experimentally demonstrate that a multi-sample averaging approach to strong augmenation not only improves generalisation but reduces the sensitivity of the model's output to data augmentation. 
  
\end{itemize}

\section{Related Work}
The problem of improving image classification performance using SSL has been extensively investigated  from the classic perspective e.g.~\cite{belkin2006manifold,belkin2004regularization,zhu2003semi,zhou2004learning,grandvalet2005semi,kim2009semi}, in which one seeks to minimise a given energy functional that exploits the assumed relationship between labelled and unlabelled data \cite{SSTheory}. However, classical approaches tended to rely on hand-crafted features that limited their performance and generalisation capabilities. With the popularisation of deep learning and its ability to learn generalisable feature representations, many techniques have incorporated neural networks to mitigate problems of generalisation. These modern state-of-the-art methods are dominated by two approaches, consistency regularisation and pseudo-labelling, which differ in how they incorporate unlabelled data into the loss function. 

\subsection{Consistency Regularisation Techniques} 

One of the fundamental assumptions that allows semi-supervised learning to help performance is the \textit{cluster assumption}, which states that points in the same cluster are likely to be in the same class. This can be seen to be equivalent to the \textit{low-density assumption} which states that the decision boundaries of the model should lie in low-density regions of the data distribution.  Following from the above assumptions, if we have access to some labelled data $Z_l = \{x_i,y_i\}_{i=1}^{n_l}$ and a large amount of unlabelled data $Z_u = \{ x_i \}_{i=n_l+1}^{n_l +n_u}$, we should seek to move our decisions boundaries to be in low density regions of the joint labelled and unlabelled data distributions.

Consistency regularisation seeks to implement the low-density assumption by encouraging the model $f$ to be invariant to perturbations $\delta$ to the data $x$. As a result the decision boundaries are pushed to low-density regions. Mathematically, given some data perturbing function $u: \mathcal{X} \rightarrow \mathcal{X}$, such that $u(x) = x + \delta$, consistency based approaches seek to minimise some consistency loss $L_{\text{con}}$ in the general form of 

\begin{equation}
    L_{\text{con}} = || f(u(x)) - f(x)  ||_{2}^{2}.
\end{equation}

A large number of papers have applied this idea to SSL including the $\prod-$Model and temporal ensembling~\cite{laine2016temporal}, Virtual Adversarial Training (VAT)~\cite{miyato2018virtual}, Mean Teacher~\cite{tarvainen2017mean}, the Interpolation Consistency Training (ICT)~\cite{verma2019interpolation} RemixMatch~\cite{remixmatch} and MixMatch~\cite{berthelot2019mixmatch}. The downside of consistency regularisation techniques is the vagueness in choosing an appropriate $\delta$. This vagueness is reflected in the wide range of perturbations which have been used in the field. Virtual Adversarial Training uses adversarial training to learn an effective $\delta$ for each point. Mean Teacher \cite{tarvainen2017mean} decided to apply a perturbation to the model itself, and replaces $f(u(x))$ with an exponential moving average of the model $f_{\text{EMA}}(x)$. Interpolation Consistency Training \cite{verma2019interpolation} seeks to train the model to provide consistent predictions at interpolations of unlabelled points. The authors of~\cite{xie2019unsupervised} demonstrated that by replacing simple noising perturbations with stronger augmentation perturbations (eg, RandAugment~\cite{cubuk2020randaugment} or CTAugment \cite{remixmatch}) leads to a substantial performance improvements. Furthering on from this work, the authors of \cite{gong2021alphamatch} proposed optimisation improvements for consistency methods using alpha-divergenes and an expectation minimisation like algorithm.

\begin{figure}[t]
  \centering
  \subfloat[ \footnotesize Fully labelled data]{\includegraphics[width=0.22\textwidth]{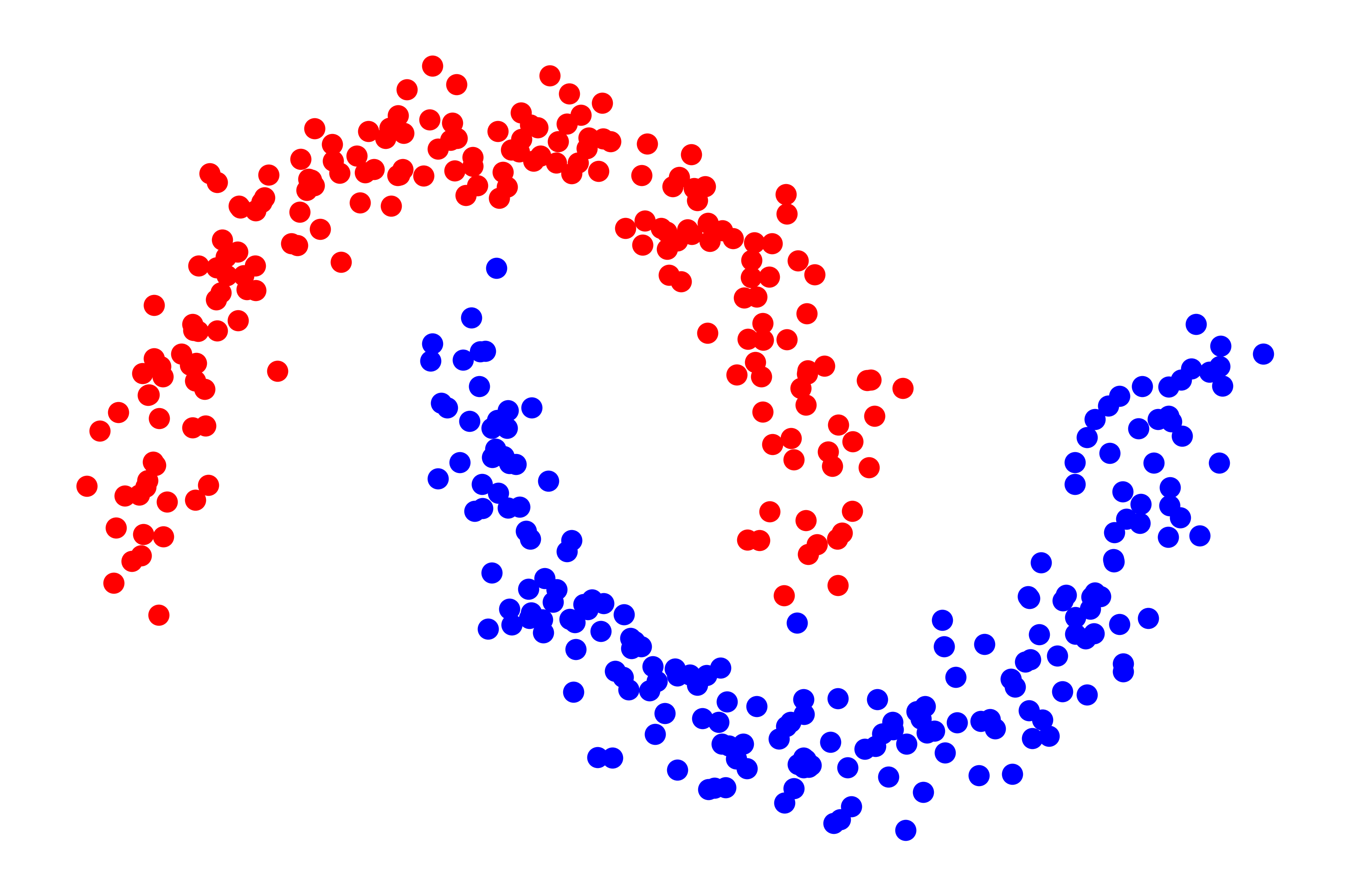}\label{fig:f1}}
  \subfloat[ \footnotesize Supervised learning]{\includegraphics[width=0.22\textwidth]{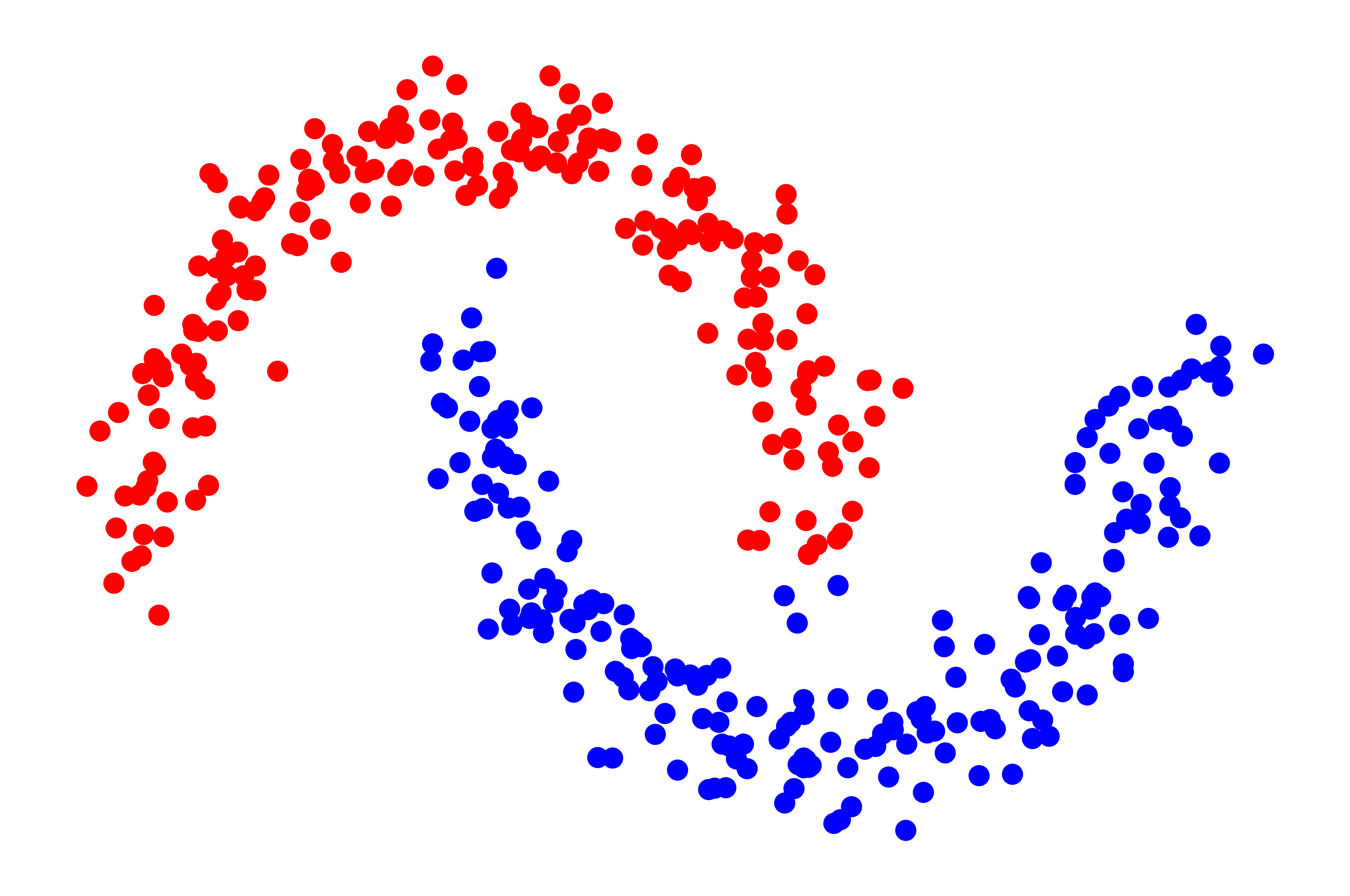}\label{fig:f2}}
  \hfill
  \subfloat[ \footnotesize Network pseudo-labels]{\includegraphics[width=0.22\textwidth]{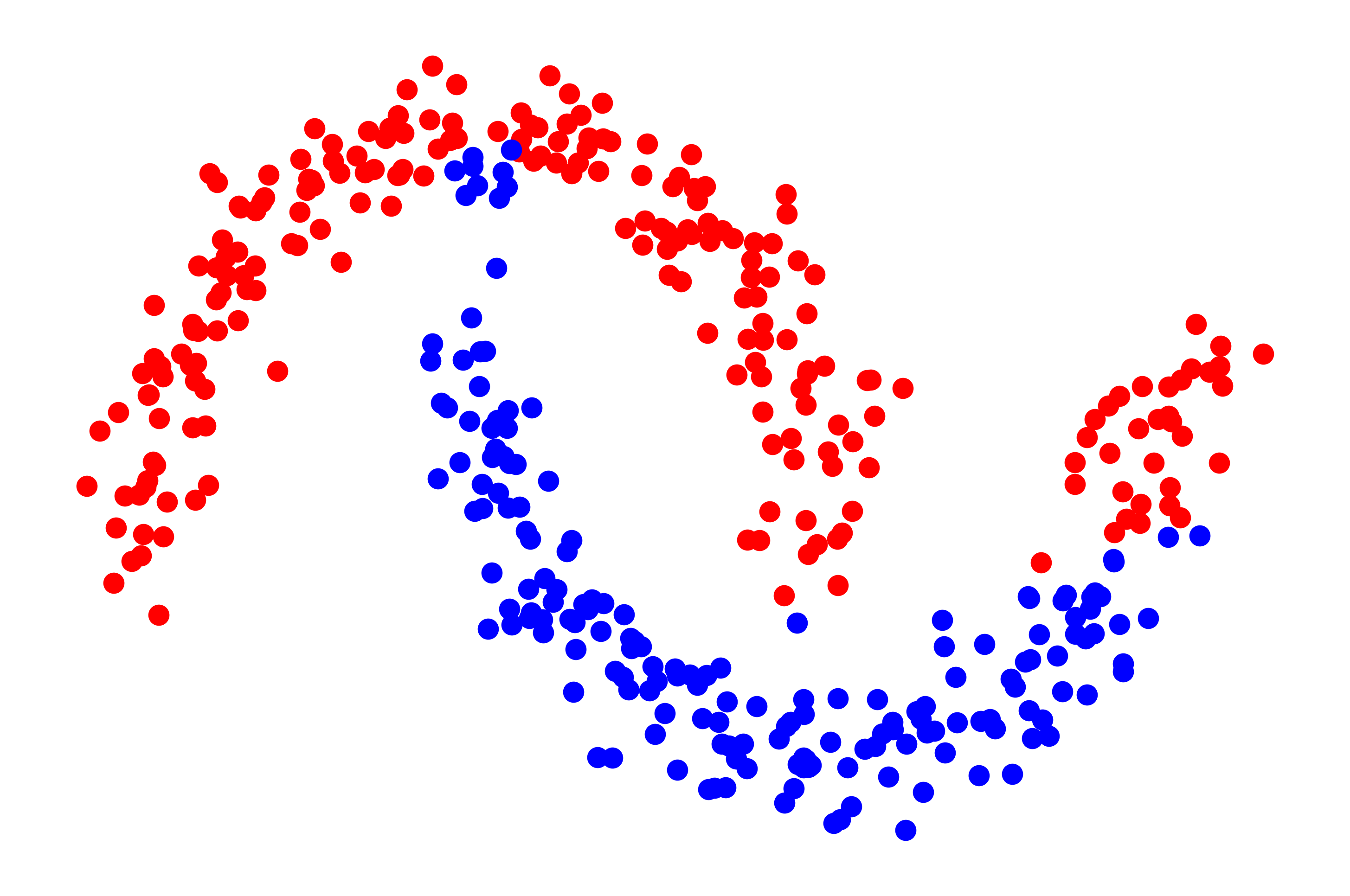}\label{fig:f3}}
  \subfloat[ \footnotesize Graphical pseudo-labels]{\includegraphics[width=0.22\textwidth]{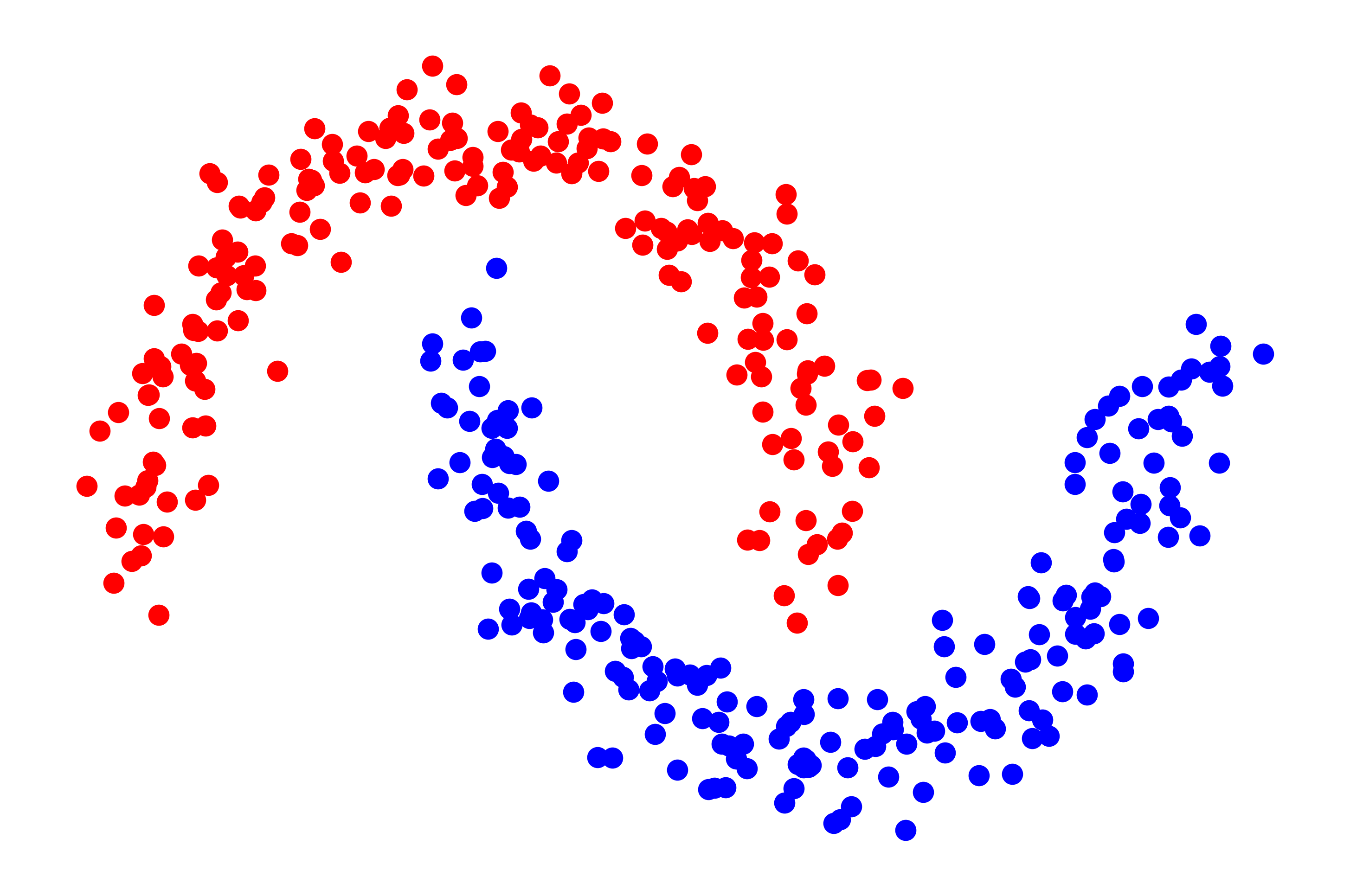}\label{fig:f4}}
  \caption{The advantage of graphical pseudo-labels on the toy \textit{Two Moon} dataset (400 samples). In (a) we show the fully labelled dataset. In (b) we show the result of supervised learning on a 80/20 train-test data split on a shallow neural network with one hidden layer. In (c) and (d) we show the output of differing semi-supervised approaches where for each we take the same shallow network but only keep 10 labeled samples per class so that $90\%$ of the data is now unlabelled. In (c) we use a pseudo-labelling approach where the pseudo-labels are generated using the common network based approach. In (d) we generate the pseudo-labels via graphical propagation. }
\label{fig:comparedmoons}
\end{figure}

Although these techniques have demonstrated great performance, it is unclear how best to set the perturbations $\delta$ and how best to incorporated them in learning frameworks. In our work, we avoid using model based perturbations and instead focus on the the application of strong data augmentation.

\smallskip
\subsection{Pseudo-Labelling Techniques} 
Another family of methods, termed pseudo-label approaches, focus on estimating labels for the unlabelled data points and then using them in a modified loss function. Forcing the network to make predictions on unlabelled points minimises the entropy of the unlabelled predictions \cite{SSTheory} and moves the decision boundaries to low-density regions. Additional, dependent on the accuracy of the pseudo-labels, we increase the amount of labelled data the model has access to and reduce overfitting to the initally small label set.   There are many ways to incorporate unlabelled data / pseudo-label pairs into the loss function but the most common ways are to either create a specific loss term for the unlabelled data pseudo-label pairs \Angie{\cite{sohn2020fixmatch,simplepseudo,xie2021muscle,assran2021semi}} or by using composite batches containing both labelled and unlabelled data and keeping the standard supervised classification loss \cite{iscen2019label,arazo2019pseudo}.

The first application of this idea to the deep learning setting was presented by Lee \cite{lee2013pseudo}. Viewing the output of the neural network $f(x)$ as a discrete probability distribution, Lee assigned a hard pseudo-label $\hat{y}$ for each unlabelled data point according to its most likely prediction $\hat{y}_i = \argmax f(x_i)$. These pseudo-labels were then used in a two termed loss function of the form 

\vspace{-0.4cm}

\begin{equation}
   \begin{aligned}
        \text{ \color{black} \textit{labelled loss} \quad \quad \quad  \color{black} \textit{unlabelled loss}} \\ 
        \hat{L}_{ssl} = \color{black} \frac{1}{n_l} \sum_{i=1}^{n_l} l_s(f(x),y) + \color{black} \eta \color{black} \frac{1}{n_u} \sum_{i=1}^{n_u} l_s(f(x),\hat{y}), \\
    \end{aligned} 
\end{equation}
\noindent
where $l_s$ is some loss function and $\eta$ is a weighting parameter. The pseudo-labels are recalculated every-time the unlabeled data is passed through the network. As an alternative to hard labels, \cite{berthelot2019mixmatch} used the full output probability distribution of the network as a soft label for each point. However, it was found that sharpening this distribution helped ensured the model's prediction entropy was minimised. 

As pointed out by Arazo et al \cite{arazo2019pseudo} there is a potential pitfall in this style of approach. Networks are often wrong and the neural network can overfit to its own incorrectly guessed pseudo-labels in a process termed confirmation bias. Arazo et al proposed using MixUp \cite{zhang2017mixup}, soft labels and a minimum ratio of labeled to unlabeled data to reduce confirmation bias. An alternative method to reduce confirmation bias is to use uncertainty quantification for the produced pseudolabels. These methods calculate a confidence score $r_i$ for each pseudo-label~$\hat{y}_i$. The works of  \cite{sohn2020fixmatch,iscen2019label,hu2021simple} used the entropy of the probability distribution to give $r_i$ whilst \cite{shi2018transductive} used the distance between unlabelled points and labelled points in feature space. One can then either weight the loss terms by $r_i$ or exclude pseudo-labels whose $r_i$ is below some threshold $\tau$ in an attempt to prevent the network learning from low confidence predictions.The work of \cite{zhang2021flexmatch} sought to improve upon a fixed threshold value by having a separate threshold for each class. The threshold was lower for classes with a greater learning difficulty, according to the model's entropy scores.

This style of approach is based upon the idea that the neural network is well calibrated, i.e that the model's softmax score is a good indicator of the actually likelihood of a correct prediction. However, recent research has suggested that modern neural networks are not as well calibrated as may be intuitively thought \cite{calibration}. In our work we demonstrate that, whilst a intuitive solution, uncertainty quantification is not needed for our pseudo-label approach. 

In a completely different direction to network predictions, it has been shown from a classical perspective \cite{zhu2003semi} that energy based models such as graphs are well suited to the task of label propagation.  Therefore, several works \cite{iscen2019label,aviles2019beyond,li2020density}  have shown good performance by iteratively feeding the feature representation of a neural network to a graph, performing pseudo-label generation on the graph and then using those labels to train the network. However, graphical approaches have yet to show that they can produce state-of-the-art results compared to model based approaches such as \cite{sohn2020fixmatch,xie2019unsupervised}.  In our work, we present a graphical approach which surpasses the performance of model based approaches, demonstrating that graphical approaches have a lot of promise for practical applications.In Fig.~\ref{fig:comparedmoons}, we display the advantage of graphical pseudo-labels using as a case study, the two moon dataset. From that figure, we observe that the graphical approach offers a clear advantage over the network pseudo-label approach.

\section{Proposed Technique}
This section details our proposed semi-supervised method. We cover the generation of pseudo-labels, the optimisation of the model alongside a full algorithm and we explore our multi-sample augmentation approach.

\smallskip
\subsection{Problem Statement:} From a joint distribution $\mathcal{Z} = (\mathcal{X},\mathcal{Y})$ we have a dataset $Z$ of size $n = n_l + n_u$ comprised of a labelled part of joint samples $Z_l= \{ x_i, y_i \}_{i=1}^{n_l}$ and a unlabelled part $Z_u = \{ x_i \}_{i=n_l+1}^{n}$ of single samples on $\mathcal{X}$. The labels come from a discrete set of size C $y \in \{1,2,..,C\}$. Our task is to train a classifier $f$, modelled by a neural network with parameter vector $\theta$, which can accurately predict the labels of unseen data samples from the same distribution $\mathcal{X}$.  The classifier $f$  can be viewed as the composition of two functions $z$ and $g$ such that $f(x) = g(z(x))$. $z : \mathcal{X} \rightarrow \mathbb{R}^{d_p}$ is the embedding function mapping our data input to some $d_p$ dimensional feature space and $g : \mathbb{R}^{d_p} \rightarrow \mathbb{R}^{C}$ projects from the feature space to the classification space.

\subsection{Pseudo-label Generation}
As a pseudo-label based approach, we iteratively assign a pseudo-label $\hat{y}$ to all data points in $Z_u$ once  per epoch. In this work, we generate hard pseudo-labels using a graph based approach first proposed by Zhou et al \cite{zhou2004learning} and first adapted to deep networks by Iscen et al \cite{iscen2019label} which has been thoroughly studied in the classical machine learning literature. We give a visual overview of our approach in Fig \ref{fig::plv}. 

We first extract the feature representation of the dataset $V$ by using the embedding function of the neural network $z$ so that $V =  \{z(x_1),..,z(x_{n}) \}$. Unlike other works we do not apply augmentation to the data whilst producing the pseudo-labels. Using $V$ and a similarity metric $d$, we use $d(v_i,v_j) = \langle v_i,v_j \rangle$, we construct a symmetric weighted adjacency matrix $W \in \mathbb{R}^{n \times n}$. The elements $w_{ij} \in W$ are given by $W_{ij} = d(v_i,v_j)$ and represent the pairwise similarities between data points. We then sparsify $W$ using the following nearest neighbour approach, which reads:

\begin{equation}
  W_{ij}=\begin{cases}
    d(v_i , v_j), & \text{if $i$ is one of the $k$ nearest neighbor of $j$,}\\
    & \text{or vice versa.}\\
    0 & \text{otherwise}.
  \end{cases}
\end{equation}

We then construct the degree matrix $D := \text{diag}(W \mathbbm{1}_n)$ and use this to normalise the affinity matrix $\mathcal{W} = D^{-1/2} W D^{-1/2}$, which prevent nodes with high degree having a large global impact. Finally, we use the initial label information to create the labelled matrix $Y \in \mathbb{R}^{n \times C}$ 

\begin{equation}
  Y_{ij}=\begin{cases}
    1, & \text{if $y_i =j$,}\\
    0 & \text{otherwise}.
  \end{cases}
\end{equation}

We can then  propagate the information contained in $Y$ across the graph structure $\mathcal{W}$ by minimising the graphical Laplacian of the prediction matrix $F \in \mathbb{R}^{n \times C}$ plus a fidelity term to the supplied labelled data:

{\small
\begin{equation}
    \mathcal{Q}(F) = \frac{1}{2}   \sum_{i,j =1}^{n} \mathcal{W}_{ij} \left|\left| \frac{F_i}{\sqrt{D_{ii}}} - \frac{F_j}{\sqrt{D_{jj}}} \right|\right|^2
    + \frac{\mu}{2} \sum_{i=1}^{n} || F_i - Y_i ||^{2} ,
    \label{graph:cost}
\end{equation}
}
where $\mu$ is a scalar weight. The first term enforces points which are close according to the metric $d$ to share a similar label whilst the second term encourages initially labelled points to keep their label. To side-step the computationally infeasible closed form solution, we use the conjugate gradient approach to solve the linear system $\left( I - \gamma \mathcal{W} \right) F = Y$, where $\gamma(1+\mu)=1$. Using $F$ the pseudo-labels $\hat{y}_i$ are given by 
\begin{equation}
    \hat{y}_i = \argmax_j F_{ij}.
    \label{eq:graph}
\end{equation}

A common problem in label propagation is that the psuedo-labels produced by the graph may be unbalanced over the classes and Iscen et al ~\cite{iscen2019label} attempted to weight the optimisation problem to avoid this possibility. We found that the weighting approach of Iscen et al actually made the performance of the model worse than leaving the predictions as is. An alternate approach to counter class in-balances is distribution alignment \cite{remixmatch}, which enforces the distribution of the pseudo-label predictions to match some given prior distribution.  The implementation of this idea by ReMixMatch focused on applying this idea to the network predictions and wasn't optimal for a graph based framework.

Instead we propose a novel smoother version of distribution alignment which can be applied during or just after the conjugate gradient approach. We give a full algorithm for this in Algorithm \ref{alg:da}. The algorithm is an iterative approach which smoothly deforms the pseudo-label predictions $F$ by the ratio $R$ between the prior distribution $D$ and the pseudo-label distribution of the unlabelled points $D_U$. Thereby promoting the prediction of underrepresented classes and vice versa. To ensure the deformation is smooth we clip the range of $R$ values to be close to one. We show in the experimental section that this approach improves the performance of the model.  

\begin{algorithm}[t!]
\caption{Smooth Distribution Alignment}
\label{alg:da}
\begin{algorithmic}[1]
    \STATE {\bfseries Input:} Pseudo-label Prediction $F \in\mathbb{R}^{n \times C}$, Prior Distribution $D \in \mathbb{R}^{C}$, labelled and unlabelled indexes $L=\{l_i\}_{i=1}^{n_l}$ and $U=\{u_i\}_{i=1}^{n_u}$ and max iteration $T$  \\
    \STATE {\bfseries Output:} Adjusted Pseudo-label Prediction $F \in\mathbb{R}^{n \times C}$  \\
    \FOR{$t_i = 1$, $t_i{+}{+}$, while $t_i < T$}
    \STATE $D_U \in \mathbb{R}^C \leftarrow \text{Initialise with zeros}$ \\
    \textbf{\textcolor{greenPython}{Get the pseudo-label distribution:}}\\
    \FOR{$u_i \in U$}
        \STATE $D_U[\argmax_j F[u_i]] \mathrel{+}= \frac{1}{n_u}$ \\  
    \ENDFOR 
    \STATE $R = D / D_u$ \\
    \textbf{\textcolor{greenPython}{ Clip values for smooth deformation:}}
    \STATE $R[R>1.01]=1.01$ and $R[R<0.99]=0.99$
    \textbf{\textcolor{greenPython}{\# Deform the current predictions:}}
    \FOR{$c_i = 1$, $c_i{+}{+}$, while $c_i < C$}
        \STATE $F[U,c_i] \mathrel{*}= R[c_i]$ 
    \ENDFOR
    \STATE Row normalise $F$ to give valid distributions. \\
    \ENDFOR
\end{algorithmic}
\end{algorithm}

\subsection{Semi-Supervised Loss}

In the deep semi-supervising setting, particularly in the current SOTA \cite{sohn2020fixmatch} \cite{berthelot2019mixmatch}, several works seek to minimise a semi-supervised loss $\hat{L}_{ssl}$ composed of two or more terms, one each for the labelled and unlabelled data points and potentially others covering entropy minimisation etc., which has the following form:
\begin{equation}
   \begin{aligned}
        \text{ \color{black} \quad \quad \quad \quad \textit{labelled loss} \quad \quad \quad  \color{black} \textit{unlabelled loss}  \quad \textit{other terms} } \\ 
        \hat{L}_{ssl} = \color{black} \frac{1}{n_l} \sum_{i=1}^{n_l} l_s(f(x),y) + \color{black} \eta \color{black} \frac{1}{n_u} \sum_{i=1}^{n_u} l_s(f(x),\hat{y}) + ..... , \\
    \end{aligned} 
\end{equation}

\noindent
where $\eta$ is a balancing parameter. For our approach we wanted to strip away as much complexity from the loss function as possible in an effort to see what elements are required for good performance.  We move away from using a  composite loss and instead only use the standard supervised loss which has worked so well in supervised image classification. To include our unlabelled data we use composite batches of size $b$ which are made up of $b_l$ labelled samples and $b_u$ unlabelled samples to which we have assigned a pseudo-label $\hat{y}$. Our semi-supervised loss, $L_{ssl}$, is given by:

\begin{equation}
    L_{ssl} = \frac{1}{b} \sum_{i=1}^{b} l_s(f(x_i),y_i).
    \label{ssl-cost}
\end{equation}


Note that in \eqref{ssl-cost} we use $y_i$ to refer to both ground truth labels and pseudo-labels for brevity. What is remarkable about this loss is its simplicity. There is no confidence thresholding of the pseudo-labels, additional weighting parameters, no consistency based terms or other regularisations.  Instead we rely upon the strength of the combination of an a energy based graphical approach to pseudo-labels estimation and the clever use of strong augmentation to increase generalisation.

\subsection{Training the model}

For initialisation purposes, we quickly extract some baseline knowledge from the dataset by minimising a supervised loss $L_{sup}$, for one hundred passes through $Z_l$. This supervised loss reads:

\begin{equation}
    L_{\text{sup}} = \frac{1}{b} \sum_{i=1}^{b} l_s( f(x_i), y_i),
    \label{sup-cost}
\end{equation}
where $b$ is the batch size and $l_s$ is the cross entropy loss. We emphasis that \eqref{sup-cost} uses only the tiny labelled set $Z_l$, and is performed once before the main semi-supervised optimisation. We then begin our main learning loop which alternates between updating the pseudo-label predictions and minimising the semi-supervised loss $L_{ssl}$ for one epoch, where we define one epoch to be one pass through the unlabelled data $Z_u$. This cycle then runs for a total of $S$ optimisation steps and the fully trained model is then tested on the relevant testing set. Note that we do use Mixup \cite{zhang2017mixup} on both $L_{sup}$ and $L_{ssl}$ with a beta distribution parameters $\alpha$. In Algorithm \ref{alg:example}, we give an overview of training our model for $S$ optimisation steps.

\begin{algorithm}[t!]
\caption{Training Scheme for LaplaceNet}
\label{alg:example}
\begin{algorithmic}[1]
   \STATE {\bfseries Input:} labelled data $Z_l= \{ x_i, y_i \}_{i=1}^{n_l}$, unlabelled data $Z_u = \{ x_i \}_{i=n_l+1}^{n}$, untrained model $f$ with trainable parameters $\theta$ and embedding function $z$. Hyper-parameters: Number of optimisation steps $S$ \\
   \textcolor{greenPython}{\# Initialisation:}
   \FOR{$i=1$ {\bfseries to} $100$}
    \STATE optimise $L_{sup}$ over $Z_l$
   \ENDFOR \\
   \STATE Set current step to zero $s_i =0$ \\
   \textcolor{greenPython}{\# Main Optimisation Process:}
   \WHILE{$s_i < S$}
   \STATE Extract features: $V = \{z(x_i)\}_{i=1}^{n}$
   \STATE Construct Graph Matrix $W$
   \STATE Degree Normalisation $\mathcal{W} = D^{\frac{-1}{2}}WD^{\frac{-1}{2}} $
   \STATE Propagate Information via $\mathcal{Q}(F)$
   \STATE Distributed Alignment on $F$
   \STATE $\hat{y_i} = \argmax F_i \text{  } \forall \text{  } n_l+1 \leq i \leq n$  \\
   \FOR{$i=1$ {\bfseries to} $\floor{\frac{n_u}{b_u}}$ }
        \STATE $B_L = \{ x_i,y_i \}_{i=1}^{b_l} \subset Z_l $ , $B_U = \{ x_i,\hat{y}_i \}_{i=1}^{b_u} \subset Z_u $
        \STATE Composite Batch $ B  = B_L \cup B_U$
        \STATE Optimise $L_{ssl}$ , $s_{i} = s_{i} + 1$ 
   \ENDFOR 
   \ENDWHILE
\end{algorithmic}
\end{algorithm}

\subsection{Multi-Sampling Augmentation}
Since the work of \cite{xie2019unsupervised}, several approaches have implemented the use of strong augmentations \cite{simplepseudo,remixmatch,sohn2020fixmatch} to the problem of semi-supervised learning, with each work having a different way of including augmentation to their framework. Very recent works \cite{simplepseudo,remixmatch} have begun using multiple augmented versions of the same unlabelled image. As yet there is no motivation for why this multiple sampling idea is preferable to alternatives such as larger batch sizes or running the code for more steps. In this section we offer a theoretical motivation for why multi-sampling improves generalization along with a mathematically bound on its performance gain. With this knowledge in mind we provide a simple method for including augmentation averaging into our SSL framework and demonstrate this approach increases accuracy and reduces the sensitivity of the model to data augmentation.

We view an augmentation strategy $A$ as a set $T$ of transformations $ t: \mathcal{X} \rightarrow \mathcal{X}$ and denote it as  $T=\{ t_1 , t_2 , .. , t_\delta \}$. The current standard approach, that the majority of existing techniques follow, is to simply sample $t \sim T$ once for each data point and compute some augmented loss $L_{Aug}$:

\begin{equation}
L_{\text{Aug}} = \frac{1}{n}\sum_{i=1}^{n} l_s(f(\color{black}t(x_i)\color{black}),y_i).
\end{equation}

However, we argue that such a simple implementation, might not extract the full information present in the augmentation. If we want to encourage our model output to be more resistant to data augmentations from $T$, and as a result produce a more generalisable model, we need to perform a multi-sample approach. To justify this, we consider the following loss $L_T$: 

\begin{equation}
    L_{T} = \frac{1}{n} \sum_{i=1}^{n} \mathbb{E}_{t \sim T} [ l(f(t(x_i)),y)],
    \label{aug-cost}
\end{equation}

\noindent
which measures the risk of the model over the entire augmentation set. If we want to minimise ~\eqref{aug-cost} then we must minimise the expected augmentation error over the entire transformation set for each data point $\mathbb{E}_{t \sim T} [ l(f(t(x_i)),y)]$. To see how a multi-sample approach helps us do just that we use Hoeffding's inequality which provides us with a probability bound that the sum of bounded independent random variables deviates from its expected value by more than a certain amount. 

Let $Z_1 , .. ,, Z_{n_a}$ be a sequence of i.i.d random variables. Assume that  $\mathbb{E}[Z] = \mu$ and $\mathbb{P}[a \leq Z_i \leq b] = 1$ for every $i$. Then, by Hoeffding's inequality, for any $\epsilon >0$, one has:

\begin{equation}
    \mathbb{P} \left[ \left| \frac{1}{n_a} \sum_{i=1}^{n_a} Z_i - \mu \right|  > \epsilon \right] \leq 2\text{exp}(-2n_a\epsilon^2 / (b-a)^2).
    \label{eq:Hoe}
\end{equation}

We can rewrite~\eqref{eq:Hoe} in context of the previously defined augmentation loss where we replace $Z_1 , .. ,, Z_{n_a}$ with $n_a$ samples from $T$ : $f(t_1(x)),f(t_2(x)),..,f(t_{n_a}(x))$

\begin{equation}
\begin{aligned}
\mathbb{P} \left[ \left| \frac{1}{n_a} \sum_{j=1}^{n_a} l(f(t(x_i)),y_i)  - \mathbb{E}_{t \sim T} [ l(f(t(x_i)),y_i)] \right|   > \epsilon \right] \\
 \leq 2\text{exp}(-2n_a\epsilon^2/b^2).
\end{aligned}    
\end{equation}

As we increase $n_a$, we converge in probability to the desired loss $\mathbb{E}_{t \sim T} [ l(f(t(x_i)),y)]$ for each data point. Subsequently, we should optimise to a lower of $L_T$ meaning that the model output will fluctuate less over the augmentation set $T$ for the used training data, and in the process make our model more generalisable. Furthermore, we can see that the probability is bounded by an exponent whose power is  $ \propto -n_{a}$. Therefore, as we increase $n_a$ the rate of decrease for the bound also decreases, maxing the first few samples far more important than later ones. This result explains prior behaviours reported but not reasoned in past papers such as \cite{remixmatch}. When using a $n_a$ sample the computational complexity increases as $O(n_a)$ but as there should be diminishing returns for increasing $n_a$ it should only be necessary to use $n_a$ values slightly above one. As we have shown that a multi-sample approach should offer generic performance increases for suitable $T$ we change \eqref{ssl-cost} and \eqref{sup-cost} to a multi-sample version. For \eqref{ssl-cost} this becomes 

\begin{equation}
    L_{ssl} = \frac{1}{b} \frac{1}{n_a} \sum_{i=1}^{b} \sum_{j=1}^{n_a}l_s(f(t_j(x_i)),y_i).
    \label{ssl-cost-aug}
\end{equation}

\noindent where the index $j$ represents repeated samples from $T$ and again $y_i$ refers to both ground truth and pseudo-labels.  In the ablation section, we perform a thorough experimental evaluation to test the theoretical predictions we have made in this section. 

\textbf{Augmentation Implementation} Similarly to other approaches we use two different augmentation strategies: one for labelled data and another for unlabelled data. However, we apply strong augmentations to both labelled and unlabelled data,  unlike past approaches \cite{sohn2020fixmatch} which reported divergences using this approach. For strong augmentations we make use of RandAugment~\cite{cubuk2020randaugment}, and CutOut augmentation \cite{devries2017improved}.  For completeness we list the full data transformations for labelled and unlabelled data in Table \ref{aug-table} and the implementation of RandAugment and CutOut in the supplementary material.

\begin{table}[t]
    \caption{The augmentation transformations used for labelled and unlabelled data. For normalisation we use the official channel}
    \centering
    \vspace{5pt}
    \begin{tabular}{|c|c|} 
    \hline
    \cellcolor{gray!25} \textsc{Labelled Transform} & \cellcolor{gray!25} \textsc{Unlabelled Transform}  \\ \hline \hline
    \multicolumn{2}{|c|}{Random Horizontal Flip}  \\ 
    \multicolumn{2}{|c|}{Random Crop and Pad}  \\ \hline
    RandAugment Sample & RandAugment Sample \\
    -  & RandAugment Sample \\ \hline
    \multicolumn{2}{|c|}{CutOut} \\ 
    \multicolumn{2}{|c|}{Normalisation} \\ \hline
    \end{tabular}
    \label{aug-table}
\end{table}

\section{Implementation and Evaluation}
In this section we detail the implementation of LaplaceNet, including hyper parameter values and training schemes, and the evaluation protocol we used to measure our model's performance and compare against the current state-of-the-art. 

\subsection{Dataset Description}
We use three image classification datasets: CIFAR-10 and CIFAR-100 \cite{cifardata} and Mini-ImageNet \cite{oneshotMiniImage}. Following standard protocol, we evaluate our method's performance on differing amounts of labelled data for each datasets.

(a) \textbf{CIFAR-10,CIFAR-100} Containing 50k training images and 10k test images, these datasets contain 10 and 100 classes respectively. The image size is small at $32$ by $32$ pixels.  We perform experiments using 500,1k,2k and 4k labels for CIFAR-10 and 4k and 10k labels for CIFAR-100.

(b) \textbf{Mini-ImageNet} A subset of the popular ImageNet dataset, containing 100 classes each with 500 training and 100 test images. The resolution of the images is $84 \times 84$ pixels and represents a much harder challenge than the CIFAR-10/CIFAR-100 datasets. We use 4k and 10k labels in our experiment.

\subsection{Implementation Details } 
\textbf{Architectures} For a fair comparison to older works we use the "13-CNN" architecture \cite{tarvainen2017mean} and for comparison to recent state-of-the-art works we use a WideResNet (WRN) 28-2 and a WRN-28-8 \cite{wideresidual} architecture. We additional use a ResNet-18 \cite{wang2017residual} for Mini-Imagenet. For all models we set the drop-out rate to $0$. For the "13-CNN" we add a $l_2$-normalisation layer to the embedding function. \textbf{Infrastructure} For all experiments, we use 1-2 Nvidia P100 GPUs. \textbf{Training Details}: We train with stochastic gradient descent (SGD) using Nesterov momentum $n_m$ with value $0.9$ and weight decay $\omega$ with value $0.0005$. We use an initial learning rate of $l_r = 0.3$ and use $S=250000$ optimisation steps in total. We utilise a cosine learning rate decay such that the learning rate decays to zero after 255000 steps. We do not make use of any $EMA$ model averaging. 

\textbf{Parameters} We list the parameter values used in Table~\ref{param-table}. Most parameter values are common parameter settings from the deep learning field and are not fine-tuned to our application. Being able to work with reasonably generic parameters is well suitable to the task of SSL where using fine-tuning over validation sets is often impossible in practical applications.

\begin{table}[t!]
\caption{List of hyperparameters used in the paper across the CIFAR-10/100 and Mini-Imagenet datasets.}
\label{param-table}
\vskip 0.15in
\begin{center}
\resizebox{0.4\textwidth}{!}{
\begin{tabular}{|c|c|c|c|}
\hline
\cellcolor{gray!25} \textsc{Parameter} & \cellcolor{gray!25} \textsc{CIFAR-10} & \cellcolor{gray!25} \textsc{CIFAR-100} & \cellcolor{gray!25} \textsc{Mini-ImageNet} \\ \hline \hline
$\alpha$ & 1.0 & 0.5 & 0.5\\
$\mu$ & 0.01 & 0.01 & 0.01 \\
$k$  & 50 &  50 & 50\\
$S$  & $2.5\times 10^{5}$ & $2.5\times 10^{5}$ & $2.5\times 10^{5}$ \\
$b$   & 300 & 100 & 100  \\
$b_l$  & 48 & 50 & 50 \\
$l_r$ & 0.03 & 0.03 & 0.1  \\
$n_m$ &  0.9 & 0.9 & 0.9  \\
$\omega$ & $5\times 10^{-4}$ & $5\times 10^{-4}$ & $5\times 10^{-4}$  \\
$n_a$ &  3 & 3 & 3 \\
\hline
\end{tabular}
}
\end{center}
\vskip -0.1in
\end{table}

\begin{table*}[t!]
  \centering
  \caption{Top-1 error rate on the CIFAR-10/100 datasets for our method and other methods using the 13-CNN architecture. We denote with $\dagger$ experiments we have ran.}
  \begin{tabular}{|l|cccc|cc|}
  \hline
   \multicolumn{1}{|c}{\cellcolor{gray!25} \textsc{Dataset}} & \multicolumn{4}{|c|}{\cellcolor{gray!25} \textsc{CIFAR-10}} & \multicolumn{2}{c|}{\cellcolor{gray!25} \textsc{CIFAR-100}}  \\ \hline
   \multicolumn{1}{|c|}{\cellcolor{gray!25} \textsc{Method}} & \cellcolor{gray!25} 500 & \cellcolor{gray!25} 1000 & \cellcolor{gray!25} 2000 & \cellcolor{gray!25} 4000 & \cellcolor{gray!25} 4000 & \multicolumn{1}{c|}{\cellcolor{gray!25} 10000} \\ \hline
   Supervised Baseline & 37.12 $\pm$ 0.89 & 26.60 $\pm$ 0.22 & 19.53 $\pm$ 0.12 & 14.02 $\pm$ 0.10 & 53.10 $\pm$ 0.34  & 36.59 $\pm$ 0.47  \\ \hline
    \multicolumn{7}{|c|}{\cellcolor{gray!25} \textsc{Consistency Based Approaches}}  \\ \hline
    $\Pi$-Model & -  & - & -  & 12.36 $\pm$ 0.31 & - & 39.19 $\pm$ 0.36  \\ 
    MT$\dagger$ & 27.45 $\pm$ 2.64 & 21.55 $\pm$ 1.48 & 15.73 $\pm$ 0.31 & 12.31 $\pm$ 0.20 & 45.36 $\pm$ 0.49 & 36.08 $\pm$ 0.51 \\ 
    VAT & - & - &  - & 11.36 $\pm$ 0.34  & -  & -  \\ 
    MT-LP & 24.02 $\pm$ 2.44 & 16.93 $\pm$ 0.70 & 13.22 $\pm$ 0.29 & 10.61 $\pm$ 0.28 & 43.73 $\pm$ 0.20 & 35.92 $\pm$ 0.47 \\ 
    SNTG & -- & 18.41$\pm$0.52  & 13.64$\pm$0.32  & 9.89$\pm$0.34  &  -- &   37.97$\pm$0.29 \\
    MT-fast-SWA & - & 15.58 $\pm$ 0.12 & 11.02 $\pm$ 0.12 & 9.05 $\pm$ 0.21 & -  & 34.10 $\pm$ 0.31  \\ 
    MT-ICT & - & 15.48 $\pm$ 0.78 & 9.26 $\pm$ 0.09 & 7.29 $\pm$ 0.02 & - & -  \\ 
    Dual Student & -- & 14.17$\pm$0.38 & 10.72$\pm$0.19 & 8.89$\pm$0.09 &  -- &  32.77$\pm$0.24\\\hline
    \multicolumn{7}{|c|}{\cellcolor{gray!25} \textsc{Pseudo-Labelling Approaches}}  \\ \hline
    TSSDL$\dagger$ & - & 21.13 $\pm$ 1.17 & 14.65 $\pm$ 0.33 & 10.90 $\pm$ 0.23 & - & -  \\
    LP$\dagger$ & 32.40 $\pm$ 1.80 & 22.02 $\pm$ 0.88 & 15.66 $\pm$ 0.35 & 12.69 $\pm$ 0.29 & 46.20 $\pm$ 0.76 & 38.43 $\pm$ 1.88  \\ 
    DAG & 9.30 $\pm$ 0.73 & 7.42 $\pm$ 0.41 & 7.16 $\pm$ 0.38 & 6.13 $\pm$ 0.15 &  37.38 $\pm$ 0.64 & 32.50 $\pm$ 0.21 \\
    Pseudo-Label Mixup & 8.80 $\pm$ 0.45 & 6.85 $\pm$ 0.15 &  - & 5.97 $\pm$ 0.15 &  37.55 $\pm$ 1.09 & 32.15 $\pm$ 0.50  \\ 
    LaplaceNet $\dagger$ & \textbf{5.68} $\pm$ \textbf{0.08} & \textbf{5.33} $\pm$ \textbf{0.02} & \textbf{4.99} $\pm$ \textbf{0.12} & \textbf{4.64} $\pm$ \textbf{0.07} & \textbf{31.64} $\pm$ \textbf{0.02} & \textbf{26.60} $\pm$ \textbf{0.23}  \\ \hline
  \end{tabular}
  \label{table:cifar-cifar-cnn-compare}
\end{table*}

\begin{table*}[t!]
  \centering
  \caption{Top-1 error rate for CIFAR-10/100. All methods, except MixMatch, are tested using the same code-base and use the same model code, the same optimiser (SGD) with the same optimisation parameters,  the same number of optimisation steps and the same RandAugment implementation. We denote with $\dagger$ experiments we have ran.}
  \begin{tabular}{|l|ccc|cc|}
  \hline \multicolumn{1}{|c}{\cellcolor{gray!25} \textsc{Dataset}} &  \multicolumn{3}{|c|}{\cellcolor{gray!25} \textsc{CIFAR-10}} & \multicolumn{2}{c|}{\cellcolor{gray!25} \textsc{CIFAR-100}}  \\ \hline
   \multicolumn{1}{|c|}{\cellcolor{gray!25} \textsc{Method}}  & \cellcolor{gray!25} 500 & \cellcolor{gray!25} 2000 & \cellcolor{gray!25} 4000 & \cellcolor{gray!25} 4000 & \multicolumn{1}{|c|}{\cellcolor{gray!25} \textsc{10000}} \\ \hline
    MixMatch & 9.65 $\pm$ 0.94 & 7.03 $\pm$ 0.15 & 6.34 $\pm$ 0.06 & --- & ---  \\ \hline
    
    \multicolumn{6}{|c|}{\cellcolor{gray!25}\textsc{Same Codebase}}  \\ \hline
    UDA $\dagger$ & 6.88 $\pm$ 0.74 & 5.61 $\pm$ 0.16 & 5.40 $\pm$ 0.19 & 36.19 $\pm$ 0.39 & 31.49 $\pm$ 0.19  \\ 
    FixMatch(RA) $\dagger$ & 5.92 $\pm$ 0.11 & 5.42 $\pm$ 0.11 & 5.30 $\pm$ 0.08  &  34.87 $\pm$ 0.17  &  30.89 $\pm$ 0.18  \\ 
    LaplaceNet $\dagger$ & \textbf{5.57} $\pm$ \textbf{0.60} & \textbf{4.71} $\pm$ \textbf{0.05}  & \textbf{4.35} $\pm$ \textbf{0.10} & \textbf{33.16} $\pm$ \textbf{0.22} & \textbf{27.49} $\pm$ \textbf{0.22} \\ \hline
  \end{tabular}
  \label{table:modern:cifar}
\end{table*}

\begin{table}[h]
\centering
\caption{Top-1 error rate for Mini-ImageNet. We compare against methods which have used an identical ResNet-18 architecture.}
\vspace{3pt}
\begin{tabular}{|l|cc|}
\hline
\multicolumn{1}{|c}{\cellcolor{gray!25} \textsc{Method}} & \cellcolor{gray!25} 4000 & \multicolumn{1}{c|}{\cellcolor{gray!25} 10000} \\ \hline
Supervised Baseline & 66.04 $\pm$ 0.32 & 52.89 $\pm$ 0.33\\ \hline

\multicolumn{3}{|c|}{\cellcolor{gray!25} Consistency Regularisation Methods}  \\ \hline
MT & 72.51 $\pm$ 0.22 & 57.55 $\pm$ 1.11\\
MT-LP & 72.78 $\pm$ 0.15 & 57.35 $\pm$ 1.66 \\ \hline
\multicolumn{3}{|c|}{\cellcolor{gray!25} Pseudo-Label Methods}  \\ \hline
LP & 70.29 $\pm$ 0.81 & 57.58 $\pm$ 1.47 \\ 
Pseudo-Label Mixup & 56.49 $\pm$ 0.51 & 46.08 $\pm$ 0.11 \\ 
LaplaceNet  & \textbf{46.32} $\pm$ \textbf{0.27} & \textbf{39.43} $\pm$ \textbf{0.09} \\ \hline
\end{tabular}
\label{mini-image:table}
\end{table}

\subsection{Evaluation Protocol}
We evaluate the performance of LaplaceNet on the CIFAR-10/CIFAR-100 and Mini-Imagenet datasets and compare against the current SOTA models for semi-supervised learning. For ease of comparison, we split the current SOTA into two groups.

\begin{enumerate}
    \item Methods which used the $13$-CNN architecture \cite{tarvainen2017mean}: $\Pi$-Model \cite{laine2016temporal}, Mean Teacher(MT) \cite{tarvainen2017mean}, Virtual Adversarial Training (VAT) \cite{miyato2018virtual}, Label Propagation for Deep Semi-Supervised Learning (LP) \cite{iscen2019label}, Smooth Neighbors on Teacher Graphs (SNTG) \cite{luo2018smooth}, Stochastic Weight Averaging(SWA) \cite{athiwaratkun2018there}, Interpolation Consistency Training (ICT) \cite{verma2019interpolation}, Dual Student \cite{ke2019dual},  Transductive Semi-Supervised Deep Learning(TSSDL) \cite{shi2018transductive}, Density-Aware Graphs (DAG)  \cite{li2020density} and Pseudo-Label Mixup \cite{arazo2019pseudo}. Unfortunately, due to the natural progress in the field, each paper has different implementation choices which are not standardised. Despite this, comparisons to this group are still useful as a barometer for model performance. 
    \item Recent methods which used the WRN \cite{wideresidual} (MixMatch \cite{berthelot2019mixmatch}, FixMatch (RandAugment variant) \cite{sohn2020fixmatch} and UDA~\cite{xie2019unsupervised}). To guarantee a fair comparison to these techniques, and as suggested by \cite{realisticssl}, we used a shared code-base for UDA and FixMatch which reimplemented the original baselines. Additionally we then ensured UDA and FixMatch used the same model code, the same optimiser with the same parameters, the same number of optimisation steps and the same RandAugment implementation as our approach.
\end{enumerate}

\textbf{Evaluation Protocol} For each dataset we use the official train/test partition and use the Top-1 error rate as the evaluation metric. For each result we give the mean and standard deviation over five label splits.

\section{Results and Discussion}

In this section, we discuss the experiments we performed to evaluate and compare our model against the state-of-the-art (SOTA). Additionally, we detail several ablation experiments which explore the benefits of graph-based pseudo-labels, the effect of augmentation averaging and evaluating the importance of individual components.

\subsection{Comparison to SOTA}
Firstly, we test our model on the less complex CIFAR-10 and CIFAR-100 datasets. In Table \ref{table:cifar-cifar-cnn-compare}, we compare LaplaceNet against the first group of methods using the $13$-CNN network. Our approach, by some margin, produces the best results on CIFAR-10 and CIFAR-100 and represents a new SOTA for pseudo-labels methods. We obtain a lower error rate using 500 labels than the recent work of Arazo et al \cite{arazo2019pseudo} obtain using 4000 labels. For CIFAR-100 LaplaceNet is a full $6\%$ more accurate than any other approach and the first method to achieve an error rate below $30\%$ on CIFAR-100 using 10k labels. In Table \ref{table:modern:cifar} we compare against the second group of methods using the WRN-28-2 network. LaplaceNet is again the best performing method, outperforming the works of UDA \cite{xie2019unsupervised} and FixMatch \cite{sohn2020fixmatch}. In particular we find a significant increase in performance on the more complex CIFAR-100 dataset and beat the other considered methods by more than $3\%$ with 10k labels.

To test the performance of LaplaceNet on a more complex dataset, we evaluate our model on the Mini-ImageNet dataset, which is a subset of the well known ImageNet dataset and in Table \ref{mini-image:table} we compare our results against all others methods which have used this dataset. Once again, we find our method performs very well, producing an error rate a $10\%$ and $7\%$ better than any other method on 4k and 10k labelled images respectively. Demonstrating our approach can be applied to complex problems in the field. Additionally, we are more than $20\%$ more accurate that the nearest graphical approach (LP).

To test the effect of increasing network size on our performance we also ran our model on CIFAR-10/100 using an WRN-28-8(26 million parameters) architecture and and compared that to the WRN-28-2(1.6 million parameters) architecture in Table \ref{table:cifar-bigger-compare}. Unsurprisingly, we achieved a large performance improvement using a WRN-28-8 on both CIFAR-10 and CIFAR-100, with an $2.87$ error rate on CIFAR-10 using 4k labels and an $22.11\%$ error rate on CIFAR-100 using 10k labels. 

From these results, we underline the strength  of our technique. Firstly,  we show that our proposed method has a performance advantage over a range of consistency and pseudo-label based methods, including other related graphical models, on both simpler and more complex tasks. Secondly, we also showed this performance behaviour to be true over a range of model architectures. Thus demonstrating that improving the pseudo-labels that are fed-back to the neural network by exploiting classical energy models can greatly improve the final test accuracy of the network.

\begin{table}
  \centering
  \caption{The effect on Top-1 error rate by scaling up the neural network in size from a WRN-28-2 to a WRN-28-8 on the CIFAR-10/100 datasets.}
  \scalebox{0.9}{
  \begin{tabular}{|l|cc|cc|}
  \hline \multicolumn{1}{|c}{\cellcolor{gray!25} \textsc{Dataset}} &  \multicolumn{2}{|c|}{\cellcolor{gray!25} \textsc{CIFAR-10}} & \multicolumn{2}{c|}{\cellcolor{gray!25} \textsc{CIFAR-100}}  \\ \hline
  \multicolumn{1}{|c|}{\cellcolor{gray!25} \textsc{Model}}  & \cellcolor{gray!25} 500 & \cellcolor{gray!25} 4000 &\cellcolor{gray!25}  4000 & \multicolumn{1}{c|}{\cellcolor{gray!25} \textsc{10000}} \\ \hline
  \multicolumn{1}{|c|}{WRN-28-2} & 5.57 $\pm$ 0.60 &4.35 $\pm$ 0.10 & 33.16 $\pm$ 0.22 & 27.49 $\pm$ 0.22 \\
  \multicolumn{1}{|c|}{WRN-28-8} & \textbf{3.81 $\pm$ 0.37} & \textbf{2.87 $\pm$ 0.18} & \textbf{26.61 $\pm$ 0.10} & \textbf{22.11 $\pm$ 0.23 } \\ \hline
  
  \end{tabular}}
  \label{table:cifar-bigger-compare}
\end{table}

\subsection{Graph Based Pseudo-Labels}
Many pseudo-label based techniques \cite{sohn2020fixmatch} \cite{arazo2019pseudo} have produced state-of-the-art results using pseudo-labels generated directly by the network rather than using an energy based approach such as label propagation on a constructed graph, which is computationally more complex. Therefore, in this section we examine the advantage of using a graph based approach. To test the importance of graph based pseudo-labels, we created two variants of LaplaceNet, both without distribution alignment and with $n_a=1$. 

\begin{enumerate}
    \item The pseudo-labels are generated directly from the network predictions: $\hat{y}_i = \text{argmax } f(x_i) \text{  } \forall \text{  } i > l$ 
    \item The pseudo-labels are generated from the graph, as in \eqref{eq:graph}, $\hat{y}_i = \text{argmax}_j \text{  } F_{ij} \text{  } \forall \text{  } i > l$ 
\end{enumerate}

\begin{figure*}
    \centering
    \subfloat[\centering 4k labels]{{\includegraphics[width=8cm]{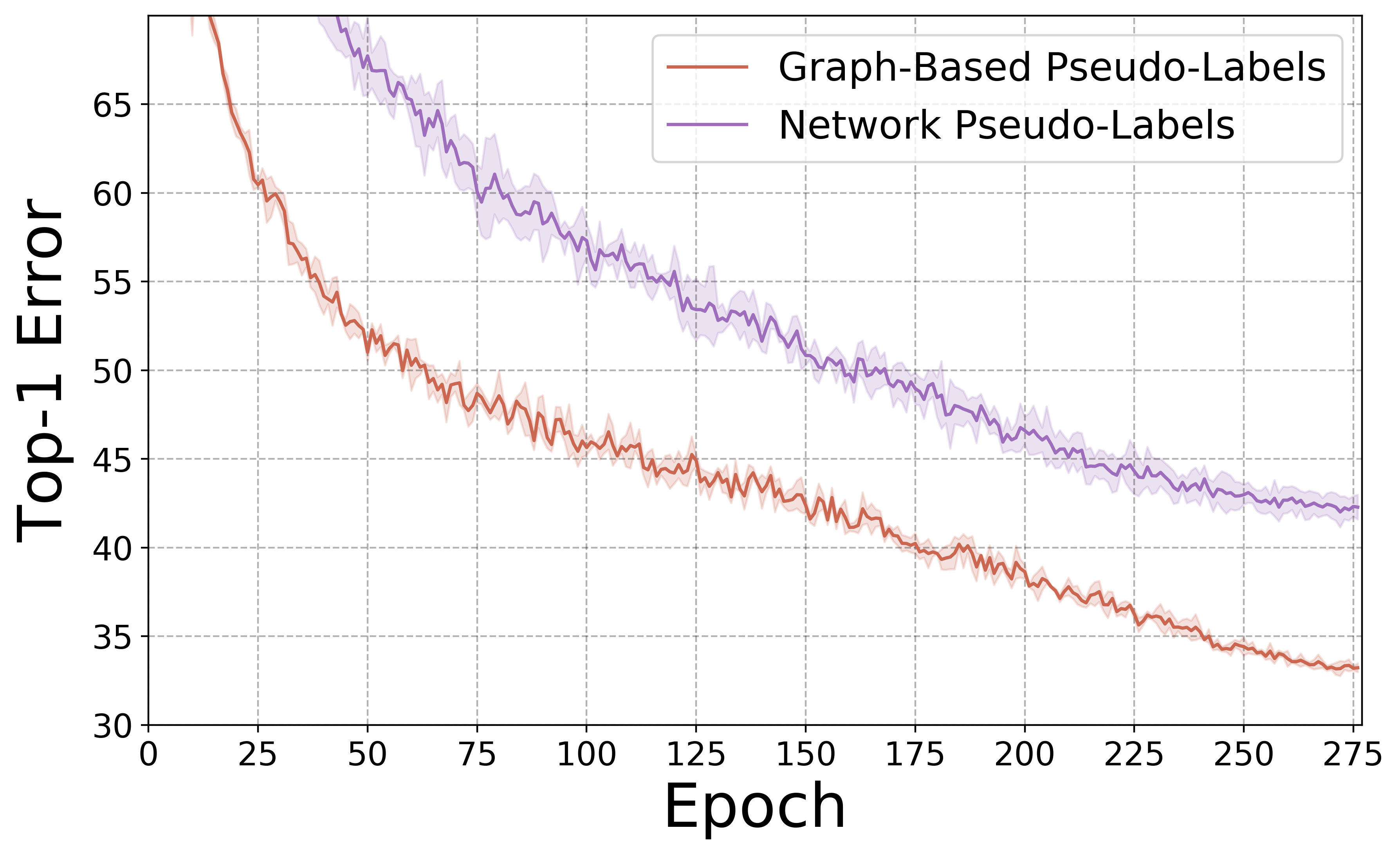} }}%
    \qquad
    \subfloat[\centering 10k labels]{{\includegraphics[width=8cm]{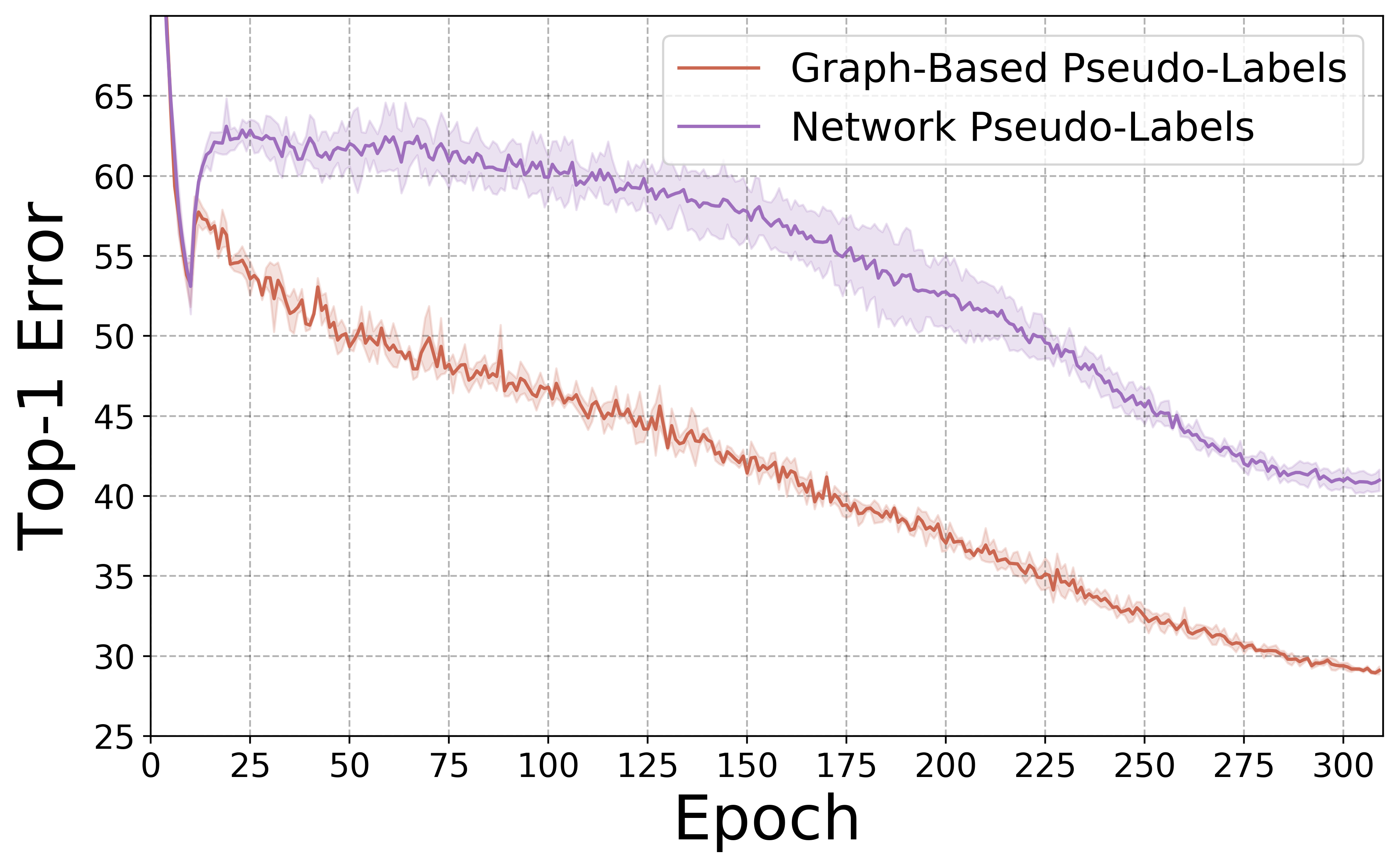} }}%
    \caption{Experimental comparison of the effect of using pseudo-labels produced in a graphical framework versus pseudo-labels generated by the neural network on the Top-1 error rate on the CIFAR-100 dataset ((a) 4k and (b) 10k labelled images) with the 13-CNN network. Using graphically produced pseudo-labels we achieve a much higher accuracy than using the network predictions.}%
    \label{fig:graphvsmodel}%
\end{figure*}

We then compared the Top-1 error rate of these two variants on the CIFAR-100 dataset, see Fig \ref{fig:graphvsmodel}. The graph-variant greatly outperformed the direct prediction variant, emphasising the clear advantage that graphically produced pseudo-labels have. What is contributing to this advantage? As an energy-based approach, propagation on the graph incorporates information on the global structure of the data, whilst the network is making a purely local decision at each point. Arazo et al \cite{arazo2019pseudo} showed that naive network based pseudo-label approach could not generate an accurate solution for the "two moons" toy dataset, despite the fact that this problem has been solved by graphical methods for some time \cite{zhu2003semi}. Thus demonstrating that purely local decisions are detrimental to accuracy when the global structure of data isn't taken into account.

\subsection{Augmentation Averaging}
In this paper we justify a multiple augmentation approach to further improve semi-supervised models. In this section, we present the experimental verification of our theoretical predictions about augmentation averaging as well as comparing its effect to potential alternative techniques. To test the effect of augmentation averaging we ran our approach on the CIFAR-100 dataset using the 13-CNN network for a range of values $n_a=[1,3,5]$. Additionally we compared the changed caused by augmentation averaging to the more common approaches of scaling the batch size $b$ and labelled batch size $b_l$ by $[1,3,5]$ and scaling the number of optimisation steps $S$ by $[1,3,5]$

To quantify the effect of a given change we use two measures: the augmentation invariance of the classifier, which we define in this paper, and Top-1 error. Augmentation invariance measures the extent to which the classifier's performance changes under data augmentation. Given an augmentation function $u : \mathcal{X} \rightarrow \mathcal{X}$ and a classifier $f_\theta$ the augmentation invariance $V$ with respect to a dataset $Z$ made up of $n$ point-label pairs $Z = \{ x_i , y_i \}_{i=1}^{n}$ is given by

\begin{equation}
    V_{Z} = \frac{\frac{1}{n} \sum_{i=1}^{n} \mathbbm{1}_{\argmax f_{\theta}(u(x_i)) = y_i}}{\frac{1}{n} \sum_{i=1}^{n} \mathbbm{1}_{\argmax f_{\theta}(x_i) = y_i}},
\end{equation}

\noindent which can be viewed as the performance ratio with and without data augmentation. We consider both the augmentation invariance of our model with respect to the fully labelled training and test data in order to give a full picture of the model's invariance, but we still only use a subset of the labelled data for training.

\begin{figure*}[t!]
    \center
      \subfloat{\includegraphics[width=0.9\textwidth]{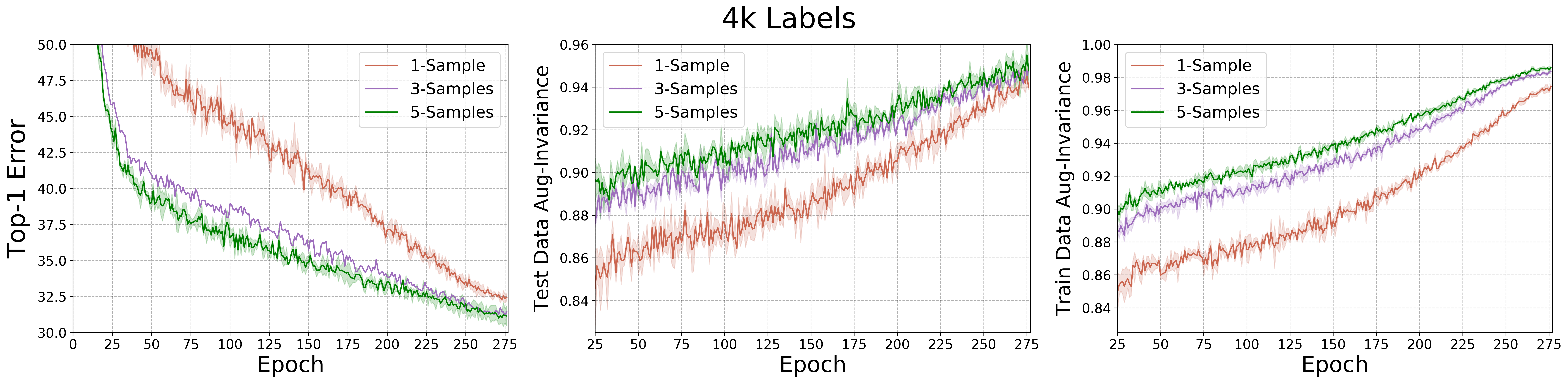}}
      \hfill
      \subfloat{\includegraphics[width=0.9\textwidth]{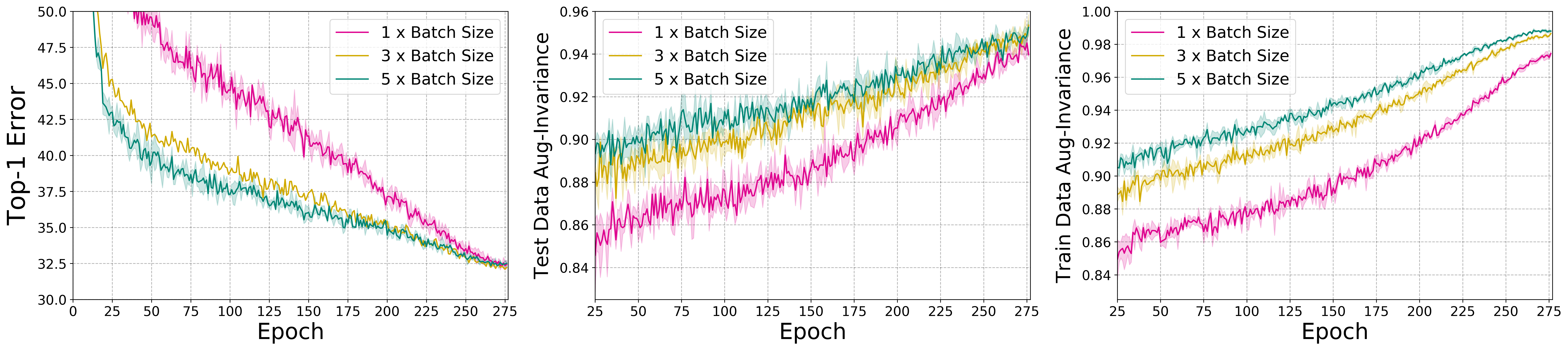}}
      \hfill
      \subfloat{\includegraphics[width=0.9\textwidth]{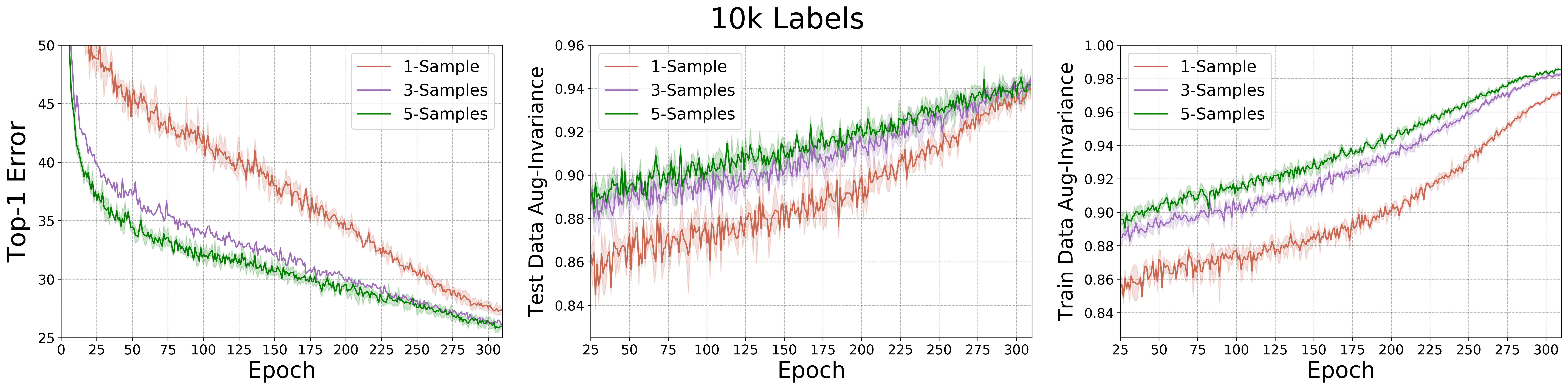}}
      \hfill
      \subfloat{\includegraphics[width=0.9\textwidth]{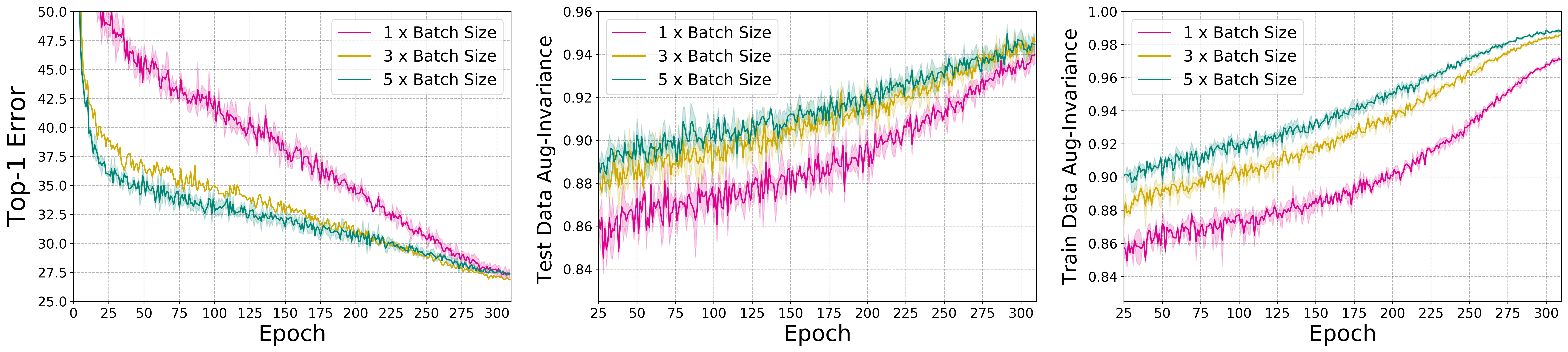}}
    \caption{A comparison on the effect of increasing batch size versus increasing the number of augmentation samples on Top-1 error rate, test data augmentation invariance and training data augmentation invariance for the CIFAR-100 dataset. Increasing the amount of augmentation averaging decreased the error rate whilst also decreasing the sensitivity of the model's output predictions to augmented data. Increasing the batch size had a similar effect on the model's sensitivity, but it offered no improvement to model accuracy.}
    \label{Fig:batchvsview}
\end{figure*}

In Fig~\ref{Fig:batchvsview} we present our findings. We found that naively scaling the number of optimisation steps without changing the hyperparameters led to the model diverging as we spent too many epochs at a high learning rate. Therefore, we provide results for the other two considered techniques which can be directly compared. As theorised in Section III we find that increasing the number of augmentation samples decreased the sensitivity of the model's predictions to augmentation on both the training and test data.An almost identical effect was found by scaling the batch size. However, the major difference between the two is their effect on Top-1 error rate. We found scaling the batch size offered no improvement to Top-1 error, in-fact the largest batch size offered the worst outcome, whilst increasing the number of augmentation samples noticeably improved the model's accuracy. Additionally as theorised in Section III, we see that the gain in performance from $n_a = 1 \rightarrow 3$ is much greater than $n_a = 3 \rightarrow 5$, supporting our statements regarding the exponential bound in probability. 

To further test the limits of augmentation averaging we ran additional experiments with $n_a = [7,10] $ with 4k labels on the CIFAR-10 dataset and present these results in Table \ref{tab:ablation-further}. We see that increasing $n_a$ above five led to small improvements in Top-1 Error and minimal changes in test data augmentation invariance, agreeing with the prior theoretical predictions.

\begin{table}[t]
    \centering
    \caption{The effect of further increasing the number of augmentation samples for each image on the CIFAR-100 dataset using the 13-CNN network and 4K labelled images.}
    \begin{tabular}{|l|cc|}
    \hline
    \cellcolor{gray!25} \textsc{Dataset} & \multicolumn{2}{|c|}{\cellcolor{gray!25} \textsc{CIFAR-100}} \\\hline
    \multicolumn{1}{|l|}{\cellcolor{gray!25} \textsc{$n_a$}} & \cellcolor{gray!25} Top-1 Error &  \multicolumn{1}{c|}{ \cellcolor{gray!25} Test-Data Aug-Invariance} \\ \hline
    \multicolumn{1}{|l|}{$n_a = 1$} & 32.44 $\pm$ 0.26  & 93.97 $\pm$ 0.01  \\ \multicolumn{1}{|l|}{$n_a = 3$} & 31.45 $\pm$ 0.18  & 94.84 $\pm$ 0.01  \\
    \multicolumn{1}{|l|}{$n_a = 5$} & 31.15 $\pm$ 0.38  & 95.11 $\pm$ 0.04  \\
    \multicolumn{1}{|l|}{$n_a = 7$} & 31.02 $\pm$ 0.28  & 95.09 $\pm$ 0.03  \\
    \multicolumn{1}{|l|}{$n_a = 10$} & 30.95 $\pm$ 0.25 & 95.18 $\pm$ 0.06  \\
    \hline
    \end{tabular}
    \label{tab:ablation-further}
\end{table}

\begin{table}[t]
    \centering
    \caption{The effect of removing individual components from the baseline model on Top-1 error rate for CIFAR-100 on the 13-CNN network.}
    \begin{tabular}{|l|cc|} 
    \hline
    \textsc{Dataset} \cellcolor{gray!25}& \multicolumn{2}{|c|}{\cellcolor{gray!25} \textsc{CIFAR-100}} \\\hline
    \multicolumn{1}{|l|}{\cellcolor{gray!25} \textsc{Model}} & \cellcolor{gray!25} 4k & \multicolumn{1}{c|}{ \cellcolor{gray!25} 10k} \\ \hline
    \multicolumn{1}{|l|}{Baseline} & 32.41 $\pm$ 0.25 & 27.37 $\pm$ 0.20  \\ \hline
    \multicolumn{1}{|l|}{\cellcolor{gray!25} \textsc{Component Removed}} & \cellcolor{gray!25} 4k & \multicolumn{1}{c|}{ \cellcolor{gray!25}  10k}  \\ \hline
    \multicolumn{1}{|l|}{RandAugment} & 44.43 $\pm$ 0.66 & 34.75 $\pm$ 0.23  \\
    \multicolumn{1}{|l|}{Distribution Alignment} & 33.26 $\pm$ 0.24 & 29.07 $\pm$ 0.07 \\
    \multicolumn{1}{|l|}{MixUp} & 33.74 $\pm$ 0.26 & 28.02 $\pm$ 0.20  \\  \hline
    \end{tabular}
    \label{tab:ablation-component}
\end{table}

Together these results suggests that scaling the number of augmentation samples could be a great option for semi-supervised models using suitable strong augmentations. 

\subsection{Component Evaluation}

As LaplaceNet combines several different techniques, we tested the importance of strong augmentation, distribution alignment and MixUp to the overall accuracy of the model. We created a baseline model ($n_a = 1$) and then remove each component one at a time and tested the performance on the CIFAR-100 dataset, see Table \ref{tab:ablation-component}. Whilst the removal of each component decreased the performance of the model, it is clear the most crucial component to model performance is strong augmentation and removing it drastically reduces model accuracy. However, unlike other works \cite{arazo2019pseudo} we find that whilst MixUp \cite{zhang2017mixup} offers a small advantage is it not critical for composite batch pseudo-label approaches. This may be due to the advantages of graph-based approaches overcoming the flaws of naive neural network predictions. 

\vspace{0.5cm}

\section{Conclusion}

We propose a new graph based pseudo-label approach for semi-supervised image classification, LaplaceNet, that outperforms the current state-of-the-art on several datasets whilst having a much lower model complexity. Our model utilises a simple single term loss function without the need for additionally complexity whilst additionally avoiding the need for confidence thresholding or temperature sharpening which was thought to be essential for state-of-the-art performance. We instead generate accurate pseudo-labels through a graph based technique with distribution alignment.  We also explore the role that augmentation plays in semi-supervised learning and justify a multi-sampling approach to augmentation which we demonstrate through rigorous experimentation improves both the generalisation of the network as well as the model's sensitivity to augmented data.

\section*{Acknowledgment}
PS thanks the UK Engineering and Physical Sciences Research Council (EPSRC) and the National Physical Laboratory (NPL) for supporting this work. AIAR gratefully acknowledges the financial support
of the CMIH and CCIMI University of Cambridge. CBS acknowledges support from the Philip Leverhulme Prize, the Royal Society Wolfson Fellowship, the EPSRC grants EP/S026045/1 and EP/T003553/1, EP/N014588/1, EP/T017961/1, the Wellcome Innovator Award RG98755, the Leverhulme Trust project Unveiling the invisible, the European Union Horizon 2020 research and innovation programme under the Marie Skodowska-Curie grant agreement No. 777826 NoMADS, the Cantab Capital Institute for the Mathematics of Information and the Alan Turing Institute.

\appendices

\vspace{-0.3cm}
\section{Augmentation Pool}
In this work we use RandAugment \cite{cubuk2020randaugment} rather than a learnt augmentation strategy such as AutoAugment \cite{cubuk2018autoaugment} which has a large computational cost. In Table \ref{tab:transformations} we detail the augmentation pool used. Additionally, we apply CutOut \cite{devries2017improved} augmentation after RandAugment sampling.

We use two different augmentation strategies in our work: one for labelled data and one for unlabelled data. We use "strong" augmentations, RandAugment and CutOut, on both labelled and unlabelled data with the only difference being that we sample once from RandAugment for labelled data and twice for unlabelled data. Given a pre-selected list of transformations, RandAugment randomly samples from the list with each transformation having a magnitude parameter. Rather than optimising this parameter on a validation set, which may not exist in semi-supervised applications, we sample a random magnitude from a pre-set range. This is same as is done in FixMatch \cite{sohn2020fixmatch} and UDA \cite{xie2019unsupervised}. We list the transformations for RandAugment and the implementation of CutOut in Table \ref{tab:transformations}.

\begin{table}[t!]
    \centering
    \caption{Computational time taken for our approach using 4k labelled images on the CIFAR-100 dataset using the 13-CNN architecture. We provide the time taken for a number of different settings used in the results section. All experiments were performed using one NVIDIA P100 GPU.}
    \begin{tabular}{|cc|} \hline
       \multicolumn{1}{|c}{\cellcolor{gray!25} \textsc{Model}}   & \multicolumn{1}{|c|}{\cellcolor{gray!25} \textsc{Computational Time (Hours)}}  \\ \hline
      \multicolumn{1}{|c}{Baseline}  & \multicolumn{1}{|c|}{\textsc{7.52 $\pm$ 0.04 }}  \\  \hline
       \multicolumn{2}{|c|}{\cellcolor{gray!25} \textsc{Component Removal}} \\ \hline
       \multicolumn{1}{|l|}{No Distribution Alignment} & 6.18 $\pm$ 0.01  \\
       \multicolumn{1}{|l|}{No Strong Augmentation} &  5.84 $\pm$ 0.03 \\ 
       \multicolumn{1}{|l|}{No Graphical Propogation} & 6.32 $\pm$ 0.01  \\ \hline
       \multicolumn{2}{|c|}{\cellcolor{gray!25}\textsc{Model Scaling}} \\ \hline
       
       \multicolumn{1}{|l|}{$3\times$-Batch-size} & 12.28 $\pm$ 0.03  \\
       \multicolumn{1}{|l|}{$5\times$-Batch-size} & 17.23 $\pm$ 0.06 \\ 
       \multicolumn{1}{|l|}{$3\times$-Samples} & 12.88 $\pm$ 0.01  \\
       \multicolumn{1}{|l|}{$5\times$-Samples} &  18.14 $\pm$ 0.11 \\ \hline

    \end{tabular}
    \label{tab:computational_time}
\end{table}

\section{Computational Time}
To give clarity on the how long our code takes to run we provide the computational run times of LaplaceNet on the CIFAR-100 dataset using the 13-CNN model for a variety of settings, see Table \ref{tab:computational_time}. Each experiment was run on one P100 NVIDIA GPU. From Table \ref{tab:computational_time}, we see that the time increased caused by increasing the batch size or increasing the number of samples is very similar. Component-wise, removing strong augmentation gives the largest decrease in computational time whilst removing the graphical propogation saved just over an hour on CIFAR-100. This represent a very small time trade off given the advantages present in using graphical pseudo-labels.

\begin{table*}
\centering
\caption{List of Transformations used in our application of RandAugment as well their description and magnitude range. Additionally, we list the CutOut transformation used at the end of RandAugment sampling.}
\resizebox{\textwidth}{!}{
\begin{tabular}{|l|l|l|}
\hline
 \cellcolor{gray!25} \textsc{Transformation} & \cellcolor{gray!25} \textsc{Description} & \cellcolor{gray!25} \textsc{Range} \\ 
 \hline
 \multicolumn{3}{|c|}{\cellcolor{gray!25} \textsc{RandAugment Transformations}} \\ \hline
 Autocontrast & Maximises the image contrast by setting the darkest (lightest)
 pixel to black (white) & -----\\
 Brightness & Adjusts the brightness of the image. where $B = 0$ returns a black image.
 image & $B \in [0.05,0.95]$\\
 Color & Adjusts the colour balance of the image. $C_l = 0$
 returns a black and white image. & $C_l \in [0.05,0.95]$\\
 Contrast & Controls the contrast of the image. $C_o = 0$ returns a gray
 image. & $C_o \in [0.05,0.95]$ \\
 Equalise & Equalises the image histogram.
 & ----- \\
 Identity &  Returns the original image. & -----\\
 Posterise & Reduces each pixel to $B$ bits. & $B \in [4,8]$ \\
 Rotate & Rotates the image by $\theta$ degrees. & $\theta \in [-30,30]$ \\
 Sharpness & Adjusts the sharpness of the image, where $S = 0$ returns a
 blurred image
 & $S \in [0.05,0.95]$\\
 Shear X & Shears the image along the horizontal axis with rate $R$. & $R \in [-0.3,0.3]$ \\
 Shear Y & Shears the image along the vertical axis with rate $R$  & $R \in [-0.3,0.3]$\\
 Solarize & Inverts all pixels above a threshold value of $T$ & $T \in [0,1]$  \\
 Translate X & Translates the image horizontally by ($\lambda$×image width) pixels. & $\lambda \in [-0.3,0.3]$ \\
 Translate Y & Translates the image vertically by ($\lambda$×image height) pixels & $\lambda \in [-0.3,0.3]$ \\ \hline
 \multicolumn{3}{|c|}{\cellcolor{gray!25} \textsc{CutOut Augmentation}} \\ \hline
 CutOut & Sets a random square patch of side-length ($L$×image width)
pixels to grey & $L \in [0,0.5]$ \\ \hline
\end{tabular}}
\label{tab:transformations}
\end{table*}


%



\bibliographystyle{IEEEtran}
\bibliography{IEEEabrv,bibfile.bib}

%
\IEEEpeerreviewmaketitle

\end{document}